\documentclass[runningheads]{llncs}
\usepackage{graphicx}
\usepackage{amsmath,amssymb} 
\usepackage{color}
\usepackage{url}
\usepackage{float}
\usepackage{caption}
\usepackage{times}
\usepackage{subfig}
\usepackage{indentfirst}
\usepackage{mathrsfs}
\usepackage{booktabs}
\usepackage{multirow}
\usepackage{multicol}
\usepackage{hyperref}
\usepackage[width=122mm,left=12mm,paperwidth=146mm,height=193mm,top=12mm,paperheight=217mm]{geometry}

\begin{document}
\pagestyle{plain}
\mainmatter
\def\ECCV18SubNumber{2965}  

\title{Generative Adversarial Talking Head: \\ Bringing Portraits to Life with a Weakly Supervised Neural Network}



\author{Hai X. Pham, Yuting Wang \& Vladimir Pavlovic}
\institute{Department of Computer Science, Rutgers University\\
			\{hxp1,yw632,vladimir\}@cs.rutgers.edu}

\maketitle

\begin{abstract}
	This paper presents \textit{Generative Adversarial Talking Head} (\textit{GATH}), 
	a novel deep generative neural network that enables fully automatic facial expression 
	synthesis of an arbitrary portrait with continuous action unit (AU) coefficients. 
	Specifically, our model directly manipulates image pixels to make the unseen subject in the
	still photo express various emotions controlled by values of facial AU coefficients, 
	while maintaining her personal characteristics, such as facial geometry, skin color and 
	hair style, as well as the original surrounding background. 
	In contrast to prior work, GATH is purely data-driven and it requires neither 
	a statistical face model nor image processing tricks to enact facial deformations.
	Additionally, our model is trained from unpaired data, where the input image, with its 
	auxiliary identity label taken from abundance of still photos in the wild, 
	and the target frame are from different persons.
	In order to effectively learn such model, we propose a novel 
	weakly supervised adversarial learning framework that consists of a generator, 
	a discriminator, a classifier and an action unit estimator.
	Our work gives rise to \textit{template-and-target-free expression editing}, 
	where still faces can be effortlessly animated with arbitrary AU coefficients 
	provided by the user.
	
\end{abstract}

\setlength{\belowdisplayskip}{0pt} \setlength{\belowdisplayshortskip}{0pt}
\setlength{\abovedisplayskip}{0pt} \setlength{\abovedisplayshortskip}{0pt}

\linespread{0.9}

\section{Introduction}
\label{sec:intro}

\begin{figure}[!ht]
	\centering
	\subfloat[source and target are of the same subject]
	{\includegraphics[width=6cm]{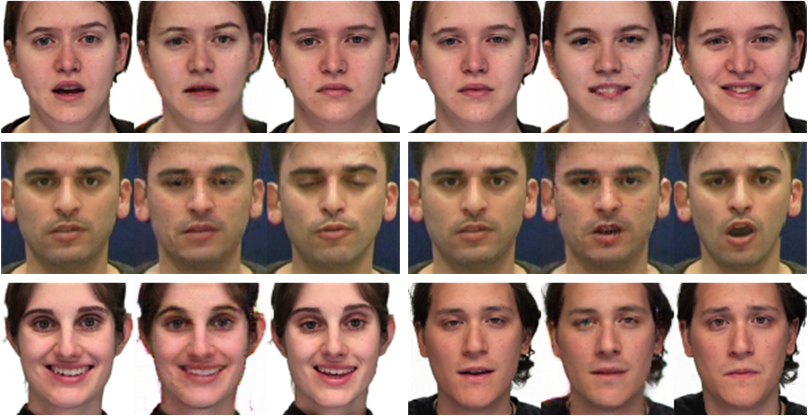}\label{fig:intro_pair}}\hfill
	\subfloat[source and target are from different persons]
	{\includegraphics[width=6cm]{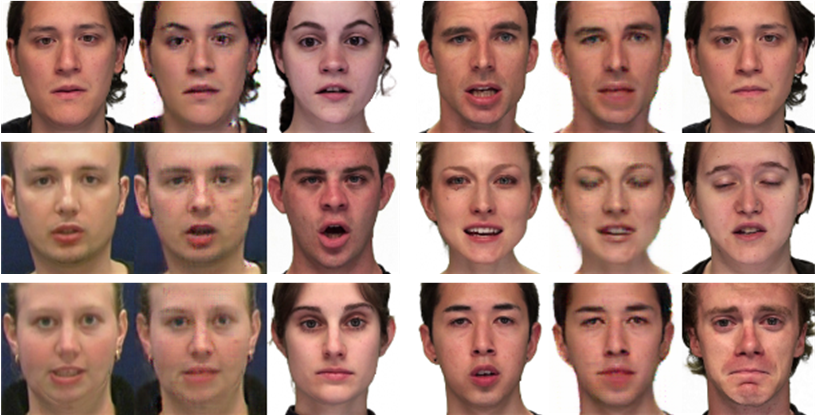}\label{fig:intro_unpair}}
	\caption{Some samples generated by our proposed GATH model.
		Each triplet consists of the source, the target and the synthesis.
		Note that our model only knows the source image and a vector of action unit
		coefficients that resemble the target.
		Could the reader tell apart which image is the source, target and synthesis?
	Hint: start with (b) first.}
	\label{fig:intro}
\end{figure}

\noindent
Human faces convey a large range of semantic meaning through facial expressions, which reflect
both actions e.g. talking, eye-blinking, and emotional states such as happy (smiling), sad (frowning)
or surprised (raising eyebrows). Over the years, much research has been dedicated to the task of
facial expression editing, in order to transfer the semantic expression from a target to a source
face, with impressive results~\cite{elor_tog17,thies_cvpr16,dale_tog11,garrido_cvpr14,olszewski_iccv17}.
In general, these state-of-the-art techniques assume that a pair of source-target images is available, 
and there exists a pair of matching 2D or 3D facial meshes in both images for texture warping and rendering.
Additionally, recent work by Thies et al.~\cite{thies_cvpr16} and Cao et al.~\cite{cao_tog16} require
a set of source images in order to learn a statistical representation, that can be used to create 
a source instance at runtime. The above requirement limits the application of these techniques to
certain settings, where source data is abundant. In~\cite{elor_tog17,olszewski_iccv17,yeh2016semantic},
the authors propose to directly transfer expressions from the target image to the source face,
forgoing the need of prior statistics of the source subject. However, there are situations in which
the target face to drive facial deformation of the source does not exist, instead, facial expression
can be inferred from other input modalities, such as speech~\cite{pham_cvprw2017,pham_arxiv17,fan_mta16},
or explicitly specified by user as vector of facial action unit (AU) intensities~\cite{facs}.

In this work, we are interested in mid-level facial expression manipulation by directly
animating a human portrait given only AU coefficients, thereby enabling a whole new level of
flexibility to the facial expression editing task. Particularly, our proposed GATH model
is able to modify a frontal face portrait of arbitrary identity and expression at pixel level,
hallucinating a novel facial image whose expressiveness mimics that of a real face that has
similar AU attributes. In other words, our model learns to extract identity features to 
preserve individual characteristic of the portrait, facial enactment to animate the portrait
according to values of AU coefficients and texture mapping, all in an end-to-end deep neural network.

\begin{figure}[!ht]
	\centering
	\includegraphics[width=\linewidth]{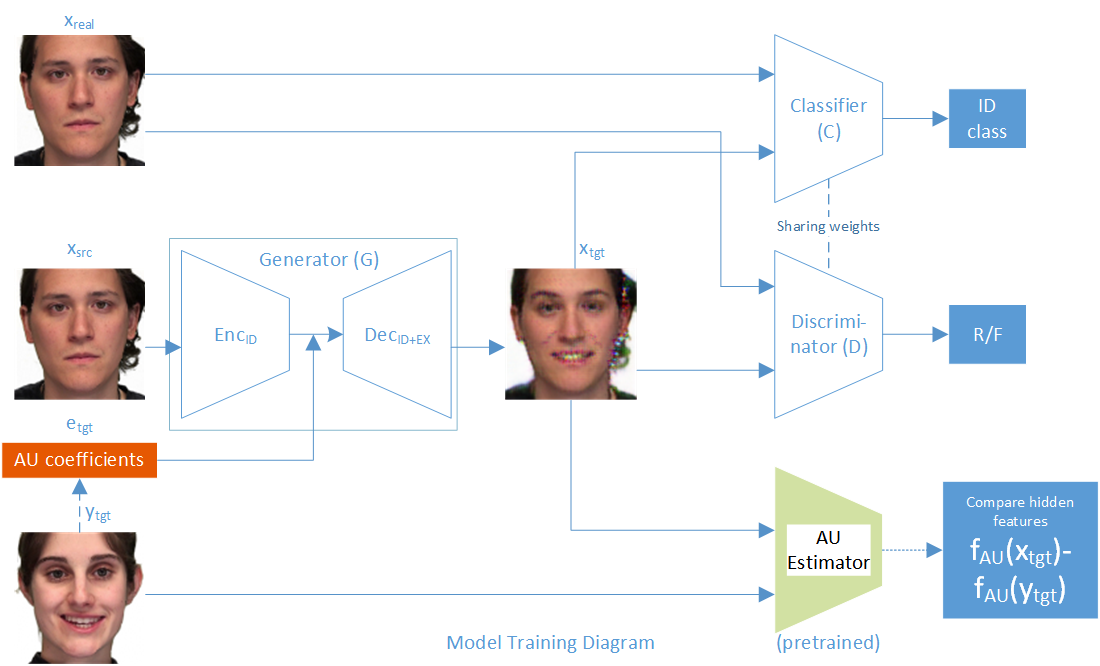}
	\caption{The proposed GATH learning framework. Except for the AU Estimator that is pretrained,
		other networks including the generator, the discriminator and the classifier 
		are jointly trained. The discriminator and the classifier share hidden layer weights.
		The generator only knows AU coefficients $e_{tgt}$ extracted from target frame $y_{tgt}$. 
		The generator $G$ learns to produce output $x_{tgt}$ from input image $x_{src}$, such that the synthesized
		face has similar facial expression as the target frame $y_{tgt}$. $x$ and $y$ are different 
		subjects.}
	\label{fig:GATH}
\end{figure}

Learning identity features requires a large number of training images from thousands of subjects,
which are readily available in various public datasets.
On the other hand, the amount of publicly available emotional videos such as~\cite{ravdess}, 
from which we could collect a wide range of AU coefficients, is rather limited. A deep net trained
on such a small number of subjects would not generalize well to unseen identity. To address this
shortcoming, we propose to train the deep net with separate source and target sets, i.e. the
animated facial image of subject A in the source set does not have an exact matching target image,
but there exists an image of subject B in the target set that has similar expression to the 
synthesized image of A, and their expressiveness similarity is measured by an auxiliary function.
Inspired by recent advances in image synthesis with adversarial learning
~\cite{goodfellow_nips14,radford_iclr16}, we jointly train the deep face generator with a
discriminator in a minimax game, in which the generator gradually improves the quality of its
synthesis to try to fool the discriminator in believing that its output is from the real facial 
image distribution. Furthermore, taking advantage of the availability of subject class labels
in the source set, we jointly train a classifier to recognize the subject label of the generated
output, therefore encouraging the generator to correctly learn identity features, and producing better
synthesis of the input subject.

Our main contributions are as follows:
\begin{itemize}
	\item Generative Adversarial Talking Head, a deep model that can generate realistic 
	expressive facial animation from arbitrary portraits and AU coefficients. 
	The model is effectively trained in an adversarial learning framework including a generator, 
	a discriminator and a classifier, where the discriminator and the classifier supervise the quality
	of synthesized images, while the generator learns facial deformations from separate source
	and target image sets, and is able to disentangle latent identity and expression code from
	the source image.
	\item An action unit estimator (AUE) network, whose hidden features are used as an expressiveness
	similarity measure between the synthetic output and its unpaired target facial image
	in order to guide the generator to synthesize images with correct expression.
	\item Extensive evaluations and applications to demonstrate the effectiveness and flexibility 
	of our proposed model in animating various human portraits from video-driven and user-defined
	AU coefficients.
\end{itemize}

\section{Related Work}
\label{sec:relwk}
\indent\textbf{"Talking head"}.
As the title suggests, our work is inspired by the line of research in synthesizing
talking faces from speech
~\cite{fan_mta16,xie_multimed07,sako_icslp00,pham_cvprw2017,karras_sigg2017,supasorn_sigg2017,taylor_sigg2017},
in which the face is animated to mimic the continuous time-varying context (i.e. talking) 
and affective states carried in the speech. 
In recent years, deep neural networks (DNN) have been successfully applied to speech synthesis ~\cite{qian_icassp14,zen_icassp13} and facial animation
~\cite{ding_mta15,zhang_interspk13,fan_mta16} with superior performance.
Taylor et al.~\cite{taylor_sigg2017} propose a system using DNN to estimate active appearance 
model (AAM) coefficients from phonemes. 
Suwajanakorn et al.~\cite{supasorn_sigg2017} use long short-term memory (LSTM) 
recurrent net to predict 2D lip landmarks from input acoustic features for lip-syncing. 
Fan et al.~\cite{fan_mta16} estimate AAM coefficients of the mouth area, 
whose texture is grafted onto an actual image to produce a photo-realistic talking head.
Pham et al.~\cite{pham_cvprw2017} map acoustic features to action unit coefficients
with an LSTM network to drive a 3D blendshape face model. Such model can
be integrated into our GATH framework to drive facial expression synthesis from speech.


\textbf{Generative Adversarial Nets (GAN)}. 
Proposed by Goodfellow et al.~\cite{goodfellow_nips14},
GAN learns the generative model in a minimax game, in which the generator and
discriminator gradually improve themselves. 
\begin{equation}
\mathop {\min }\limits_G \mathop {\max }\limits_D {E_{x \sim {p_{data}}(x)}}\left[ {\log D(x)} \right] + {E_{z \sim p(z)}}\left[ {\log \left( {1 - D(G(z))} \right)} \right].
\label{eq:gan}
\end{equation}
Eventually the generator learns to create realistic data able to fool the discriminator.
GAN has been widely used in image synthesis with various successes
~\cite{denton_nips15,ledig_cvpr15,radford_iclr16,reed_icml16,isola_cvpr17,stargan}.
Moreover, recent works also introduce additional constraints for topic-driven synthesis
~\cite{mirza_arxiv14}, or use class labels in semi-supervised GAN training
~\cite{springenberg_iclr16,odena_icml17,tran_cvpr17,li_triGan_nips17}. In our approach,
a classifier is jointly trained with the discriminator to predict synthesized images 
into $C$ classes. Consequently, not only the generator learns to generate realistic images, 
but the synthesis also preserves the identity presented in the source image.

\textbf{Facial image editing.}
Facial editing techniques in literature are mostly model-based, using a 3DMM~\cite{blanz_99,vlasic_tog05},
and follow a common approach, in which a 3DMM is fitted to both source and target images, and the target
expression is transferred to the source frame by manipulating the model coefficients to calculate
a new 3D facial shape, followed by texture remapping
~\cite{blanz_03,thies_cvpr16,garrido_cvpr14,dale_tog11,yang_tog11,yang_cvpr12}.
Averbuch-Elor et al.~\cite{elor_tog17} only use 2D face alignment with clever texture warping 
and detail transfer~\cite{liu_2001}. Instead of using graphics-based texture warping, Orszewski et al.~\cite{olszewski_iccv17} utilize supervised GAN to synthesize a new albedo in the UV
texture space, given matching source and target images. 
In a different approach, Liu et al.~\cite{liu_3dAUGAN} uses conditional GAN
to synthesize expression coefficients of a 3DMM given discrete AU labels, followed by standard
shape calculation and texture remapping. Based on variational autoencoder (VAE)~\cite{kingma_iclr13},
Yet et al.~\cite{yeh2016semantic} train an expression flow VAE from matching source-target pairs,
and edit the latent code to manipulate facial expression. \textit{Deepfakes}~\cite{deepfakes}, 
which has gone viral
on the internet recently, uses coupled autoencoders and texture mapping to be able to swap identities
of two actors.
In contrast to model-based approaches,
our model directly generates the facial image from source portrait and target AU coefficients without
using a statistical face model or a target video for transferring, forgoing
manual texture warping and blending as these tasks are automatically carried out by the deep net. 
Different from recent GAN-based synthesis models, 
our method trains the generator from totally unmatched source-target pairs.

\textbf{Representation disentangling}. 
It is still an open question of how to design proper objectives
that can effectively learn good latent representation from data. 
Kulkarni et al.~\cite{kulkarni_nips15}, Yang et al.~\cite{yang_nips15} propose models
that can explicitly separate different codes (object type, pose, lighting)
from input images. Those codes can be manipulated to generate a different looking image,
e.g. by changing the pose code. Peng et al.~\cite{peng_eccv16} propose a recurrent encoder-decoder
network for face alignment, that also learns to separate identity and expression codes
through a combination of auxiliary networks and objectives.
Tran et al.~\cite{tran_cvpr17} employ semi-supervised GAN to learn disentangled face 
identity/pose for face frontalization. GATH is similar in spirit to~\cite{tran_cvpr17},
in which the encoder subnetwork of the generator learns the latent identity code 
independently from the arbitrary expression in the source image, and the decoder takes in the combined
identity and AU code to generate an animated image from the source portrait.

\section{Proposed Model}
\label{sec:model}

\subsection{Problem Formulation}
\noindent
We first denote the following notations that will be used throughout the paper:
$G$ = generator; $f_{G\_en}$ = encoder subnetwork of $G$; $f_{G\_de}$: decoder subnetwork of $G$;
$D$ = discriminator; $\mathcal{C}$ = classifier; $\mathcal{E}$ = AU Estimator; 
$f_{au}$ = a function that maps image to a latent facial expression space;
$x_{src}$ = source portrait to be transformed;
$x_{tgt}$ = image synthesized by the generator;
$x_{re}$ = real image used to train $D$ and $\mathcal{C}$;
$c$ = the class label associated with $x_{re}$;
$y_{tgt}$ = target image;
$e_{tgt}$ = continuous AU coefficient vector corresponding to $y_{tgt}$.
$x_{src}$ and $x_{re}$ are sampled from the same training source set, and are not
necessarily the same.
$y_{tgt}$ is sampled from the training target set.
$e_{tgt}$ is a 46-D vector in which each component varies freely in [0,1],
following the convention of the FaceWarehouse database~\cite{cao_fwh2013gy}.

Our general GATH framework is illustrated in Figure~\ref{fig:GATH}.
The generator $G$ synthesizes $x_{tgt}$ from the input $x_{src}$ given AU coefficients $e_{tgt}$:
$ x_{tgt} = G(x_{src}, e_{tgt}) = f_{G\_de}(f_{G\_en}(x_{src}), e_{tgt}) $. Since $x_{src}$
may contain arbitrary expression, the generator specifically disentangles the latent identity 
code from expression features in the source image with the encoder, effectively making 
the transformation of the source face independent of the expression manifested in the input image.

Unlike previous work~\cite{olszewski_iccv17,yeh2016semantic}, our source and target sets are
disjoint\footnote{In training, we actually mix a small amount of image samples of subjects in the 
target set into the source set, which accounts for 2.7\% of its size, to increase its diversity.}.
In other words, an exact correspondence $y^0_{tgt}$ of $x_{tgt}$ does not exist, hence we are unable to
use the conventional pixel-wise reconstruction loss to learn facial deformation. However, there
exists a frame $y_{tgt}$ that shares similar values of AU intensities to $x_{tgt}$. One might naively
minimize the difference $\left\| x_{tgt} - y_{tgt} \right\|$, but using this loss has major
drawbacks: 
Firstly, there is not necessarily pixel-wise correspondence between $x_{tgt}$ and $y_{tgt}$,
hence a local facial deformation at a specific coordinate in the target does not mean that
the same visual change would also happen at the exact same coordinate in the source.
Secondly, directly minimizing the difference between the source and the target frame would
make the model learn to hallucinate the identity of the target into the source, which violates
the identity preserving aspect of our model.
Furthermore, what we want to compare is the expressiveness similarity of the source and target, 
not their entire appearances. Inspired by recent work in artistic style transfer~\cite{gatysEB15a},
we wish to compare the source and target in a latent expressiveness space with a projecting function
$f_{au}$. Thus, we propose to train a deep Action Unit Estimator network, and measure the similarity
of source and target in the hidden feature space of AUE.

One core objective of our work is to learn a generator $G$ that can generate realistic looking face 
synthesis indistinguishable from a real image, especially in our case where the exact corresponding 
target does not exist. 
To this end, we integrate the adversarial loss proposed by Goodfellow et al.~\cite{goodfellow_nips14}
into our framework, by jointly training a discriminator $D$ that can tell the difference between real
and fake images, that eventually guides $G$ to generate "fine enough" samples via a minimax game.

A straightforward approach to learn identity disentanglement is to minimize the intra-subject 
reconstruction loss $\left\| x_{tgt} - x_{src} \right\|$, as they are largely similar except
sparse local deformation parts.
However, we can utilize the available auxiliary class labels of
the source set to provide additional feedback to make the generator learn the disentanglement
more effectively. We propose to jointly train a classifier $\mathcal{C}$ that share all
hidden layer weights with the discriminator. The advantages of this approach are two-fold.
First, jointly learning the classifier $\mathcal{C}$ and discriminator $D$ helps 
discover relevant facial hidden features better, and $D$ can tell apart the real image
from the fake more easily. In return, these players provide stronger feedback to the generator,
encouraging $G$ to generate finer synthesis and better preserve the identity of the source.

\subsection{Action Unit Estimator}

\noindent
We propose a CNN model $\mathcal{E}$ based on VGG-9 architecture~\cite{Simonyan14c} to predict 
AU coefficients from a facial image: $\mathcal{E}:x \rightarrow e$. The network architecture
is shown in Figure~\ref{fig:AUE}. $\mathcal{E}$ is learned by minimizing the squared loss:

\begin{equation}
\mathop {\min }\limits_{\mathcal{E}} \left\| {{e_{gt}} - \mathcal{E}\left( x \right)} \right\|_2^2,
\label{eq:AUE}
\end{equation}
where $e_{gt}$ is a ground truth AU vector. In essence, AUE learns hidden features tailored to
facial expression and independent of identity, as well as invariant to position and scale of the face
in the image. We take the last convolutional layer of AUE as the latent space mapping function $f_{au}$
to measure the expressiveness similarity of images. First, this layer retains high-level features with
rich details to represent the facial expression. Furthermore, the convolutional layer still preserves 
the spatial 2D structure layout, hence it can pinpoint where in the source image that the local
deformation should happen.

\subsection{GATH Learning}

The models in our framework are jointly trained by optimizing the following composite loss:

\begin{equation}
{{\mathcal{L}}_{au}} + {\lambda _{rec}}{{\mathcal{L}}_{rec}} + {\lambda _{adv}}{{\mathcal{L}}_{adv}} + {\lambda _{cls}}{{\mathcal{L}}_{cls}} + {\lambda _{tv}}{{\mathcal{L}}_{tv}}.
\label{eq:L_all}
\end{equation}

The first term is the AU loss, to make $G$ learn to expressively transform the source image 
to manifest the target emotion:

\begin{equation}
{{\mathcal{L}}_{au}} = \mathop {\min }\limits_G \left\| {{f_{au}}\left( {{y_{tgt}}} \right) - {f_{au}}\left( {G\left( {{x_{src}},{e_{tgt}}} \right)} \right)} \right\|_2^2.
\label{eq:L_au}
\end{equation}

The intra-subject reconstruction loss, ${\mathcal{L}}_{rec}$, minimizes the pixel-wise difference 
between the source and the synthesized image, to preserve the subject identity as well as the background:

\begin{equation}
{{\mathcal{L}}_{rec}} = \mathop {\min }\limits_G {\left\| {{x_{src}} - G\left( {{x_{src}},{e_{tgt}}} \right)} \right\|_1}.
\label{eq:L_rec}
\end{equation}

The third term is the adversarial loss. In this work, we replace the vanilla Jensen-Shannon divergence 
GAN loss in~\eqref{eq:gan} with the least square loss proposed in CycleGAN~\cite{zhu_cyclegan_iccv17}, 
as we found that this objective makes optimization more stable.

\begin{equation}
{{\mathcal{L}}_{adv}} = \mathop {\max }\limits_G \mathop {\min }\limits_D {\left( {1 - D\left( {{x_{re}}} \right)} \right)^2} + {\left( {D\left( {G\left( {{x_{src}},{e_{tgt}}} \right)} \right)} \right)^2}.
\label{eq:L_adv}
\end{equation}

The classifier loss, ${\mathcal{L}}_{cls}$, consists of two cross-entropy loss terms.
Minimizing the first term updates the weights of the classifier, 
while the second term updates the weights of $G$.
Intuitively, the classifier updates its parameters from real image samples $x_{re}$, and provides
feedback to the generator, such that $G$ learns to generate better samples to lower the classification loss,
and consequently preserve the source identity better.

\begin{equation}
{{\mathcal{L}}_{cls}} = \mathop {\min }\limits_\mathcal{C}  - \sum\limits_i {{c_i}} \log {\mathcal{C}_i}\left( {{x_{re}}} \right) + \mathop {\min }\limits_G  - \sum\limits_i {{c_i}} \log {\mathcal{C}_i}\left( {G\left( {{x_{src}},{e_{tgt}}} \right)} \right).
\label{eq:L_cls}
\end{equation}

The last term is total variation loss to maintain spatial smoothness:

\begin{equation}
{{\mathcal{L}}_{tv}} = \sum\limits_{i,j} {{{\left( {x_{tgt}^{i,j + 1} - x_{tgt}^{i,j}} \right)}^2} + {{\left( {x_{tgt}^{i + 1,j} - x_{tgt}^{i,j}} \right)}^2}}.
\label{eq:L_tv}
\end{equation}

In the adversarial learning framework, the generator and discriminator are alternatively and iteratively updated. 
Essentially, in GATH, the joint discriminator-classifier network is updated by minimizing this loss:

\begin{equation}
\begin{gathered}
\mathop {\min }\limits_{D,\mathcal{C}} {\lambda _{adv}}\left[ {{{\left( {1 - D\left( {{x_{re}}} \right)} \right)}^2} + {{\left( {D\left( {G\left( {{x_{src}},{e_{tgt}}} \right)} \right)} \right)}^2}} \right] \\
- {\lambda _{cls}}\sum\limits_i {{c_i}} \log {\mathcal{C}_i}\left( {{x_{re}}} \right),
\end{gathered}
\label{eq:min_DC}
\end{equation}
whereas minimizing the following composite loss updates the generator:

\begin{equation}
\begin{gathered}
\mathop {\min }\limits_G {{\mathcal{L}}_{au}} + {\lambda _{rec}}{{\mathcal{L}}_{rec}} + {\lambda _{tv}}{{\mathcal{L}}_{tv}} - {\lambda _{adv}}{\left( {D\left( {G\left( {{x_{src}},{e_{tgt}}} \right)} \right)} \right)^2} \\
- {\lambda _{cls}}\sum\limits_i {{c_i}} \log {\mathcal{C}_i}\left( {G\left( {{x_{src}},{e_{tgt}}} \right)} \right).
\end{gathered}
\label{eq:min_G}
\end{equation}

\section{Implementation Details}

\subsection{Network Architecture}

\begin{figure}[!ht]
	\centering
	\subfloat[The fully convolutional generator network]
	{\includegraphics[width=\linewidth]{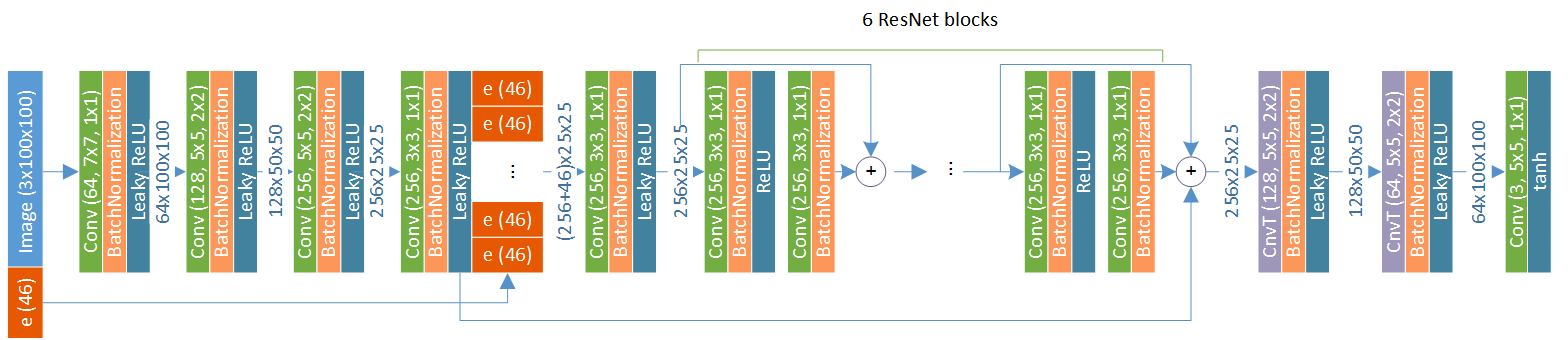}\label{fig:G}}\\
	\subfloat[The discriminator-classifier network]
	{\includegraphics[width=6cm]{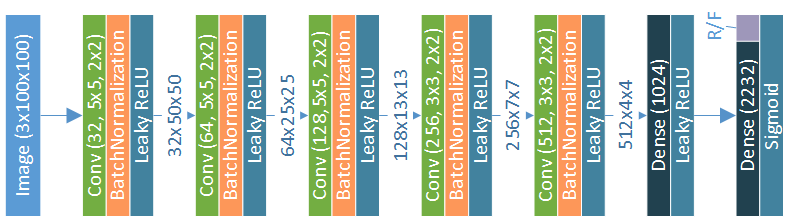}\label{fig:D}}\hfill
	\subfloat[The AU Estimator network]
	{\includegraphics[width=5.5cm]{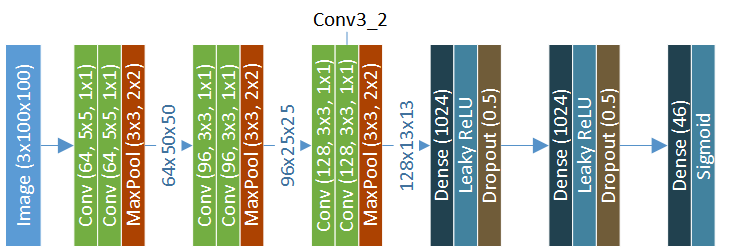}\label{fig:AUE}}\\
	\subfloat
	{\includegraphics[width=12cm]{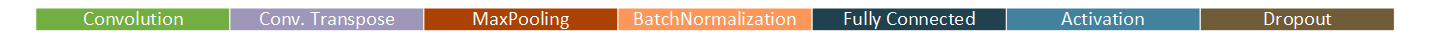}}
	\caption{\textit{(best viewed in color)}
		Architectures of deep neural networks in our GATH framework. The last
	convolutional layer of the AUE, 'conv3\_2', is used to extract expressive hidden features
	to calculate ${\mathcal{L}}_{AU}$.}
	\label{fig:GATH_arch}
\end{figure}
\noindent
Architectures of three networks in our GATH framework are illustrated in Figure~\ref{fig:GATH_arch}.
These networks are designed such that all of them can reside in the GPU memory of a Tesla K40
at the same time.

The input to each network is a standard 100x100px RGB image, with pixel values normalized to [-1,1].
The generator $G$ includes two subnetworks: encoder and decoder. The encoder consists of four
blocks, each one starts with a convolutional layer, followed by a batch normalization layer
and Leaky ReLU activation with slope factor of 0.1. 
The AU vector is concatenated to the output of the forth block by 
spatially 2D broadcasting. The decoder starts with a convolutional block, followed by six
ResNet~\cite{He2015} blocks, two convolutional transpose blocks and a final convolutional output layer.
Particularly, there is a residual connection from the output of the encoder to the end of the
bottle neck to help $G$ learn the identity code more effectively.

Figure~\ref{fig:D} shows the weight sharing discriminator-classifier network. $D$ and $\mathcal{C}$
share all hidden layer parameters, but have separate output layers.

\subsection{Training and Post-processing}

\noindent\textbf{Training.}
We organize the source set as a combination of the following still photo datasets:
Cross-age Celebrity dataset (CACD)~\cite{cacd}, FaceWareHouse~\cite{cao_fwh2013gy}, GTAV~\cite{gtav},
consisting of 2,168 identities. We also mix in a small set of frames from 20 actors in RAVDESS~\cite{ravdess} 
and 40 actors in VIDTIMIT~\cite{vidtimit} dataset, making a total of 2,228 identities in the source set.

The target set consists of a large number of frames extracted from RAVDESS and VIDTIMIT, 
two popular audiovisual datasets, in which actors display a wide range of facial actions and emotions.
We extract AU coefficients from these two datasets using a performance-driven 3D face tracker
~\cite{pham_icpr2016,pham_3dv2016}. The AUE is trained on the target set.

Furthermore, since these datasets include many face images at large pose, we use the face frontalization
technique proposed by Hassner et al.~\cite{hassner_cvpr15} to roughly convert original images into 
portraits in both source
and target sets and crop them to 100x100px. The alignment is not perfect, however, our AUE is invariant
to translation and scale, hence our generator can learn facial deformation reliably.

The models in our GATH framework, $G$, $D$ and $\mathcal{C}$, are trained end-to-end with the ADAM minibatch optimizer~\cite{adam_iclr2015}.
We set minibatch size = $64$, initial learning rate = $1e-4$ and momentum = $0.9$. Other hyperparameters in
~\eqref{eq:L_all} are empirically chosen as follows: $\lambda _{rec} = \lambda _{tv} = 1.0$,
$\lambda _{adv} = \lambda _{cls} = 0.05$. 
The whole framework\footnote{code will be published soon, stay tuned!}
 is implemented in Python based on the deep learning toolkit CNTK
\footnote{https://github.com/microsoft/cntk}.

\noindent\textbf{Post-processing.}
A side effect of using the composite loss in~\eqref{eq:L_all} to train our weakly supervised model
and pixel scaling to and from the [-1,1] range
is that the synthesis loses the dynamic range of the original input. In order to partially restore
the original contrast, we apply the adaptive histogram equalization algorithm CLAHE~\cite{clahe} to
synthesized images. Parts of evaluation where CLAHE is applied will be clearly indicated.
Furthermore, for visualization purpose, we clear the noise in the output with non-local
means denoising~\cite{ipol.2011.bcm_nlm}, followed by unsharp masking.
Figure~\ref{fig:out_syn} demonstrates the visual effects of these two enhancements on the syntheses.

\section{Evaluation}

\noindent
Due to the lack of a publicly available implementation of the face editing methods mentioned in 
Section~\ref{sec:relwk}, we simplify our GATH framework to create two baselines. 
\textbf{GATH-DC} (GATH \textit{minus} DC): GATH without the joint discriminator-classier network; 
\textbf{GATH-C}: GATH without the classifier.

For quantitative evaluation, we record facial expression synthesizing performance on a hold out 
test set of four actors from RAVDESS and three actors in VIDTIMIT, which are not included in training.
Specifically, we choose two source frames from each actor. In one frame, the actor displays
neutral (or close to neutral) expression. In the other one, the actor shows at least one
expression (mouth opening). Each source image is paired with all video sequences (94 in total),
resulting in 188 intra-class (same subject) pairs and 1032 inter-class pairs, where the source
actor is different from the target actor.

We quantify the synthesis performance of our model with a set of metrics: for intra-class
experiments, we measure the pixel-wise Mean Absolute Error (MAE) and Root Mean Square Error (RMSE), 
as well as the AU error (RMSE of intensity values), with respect to ground truth frame. 
We use the OpenFace toolkit~\cite{openface} 
to extract intensities of 17 AUs (which are different from our set of 46 AUs), each value
varies within the range of [0,5] (note that our input AU coefficients vary in [0,1]).
For inter-class experiments, we only measure the AU error.

We also provide qualitative experiments on a random set of actors from the two popular face datasets:
CelebA~\cite{liu2015faceattributes} and Labeled Face in the Wild (LFW)~\cite{LFWTech}.

Lastly, we present an application of \textit{template-and-target-free expression editing},
where user-defined AU coefficients are used to transform arbitrary sources. Mainly within the
scope of this paper, we perform expression suppression (i.e. neutralization). However, our model
is flexible enough to transform a source facial image with any arbitrary AU values.

\subsection{Intra-class Synthesis Evaluation}

\begin{figure}[!ht]
	\centering
	\null\hfill
	\subfloat[]
	{\includegraphics[width=1.25cm]{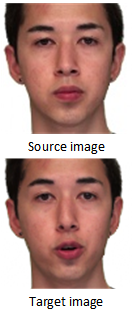}\label{fig:src_tgt}}\hfill
	\subfloat[syntheses]
	{\includegraphics[width=3cm]{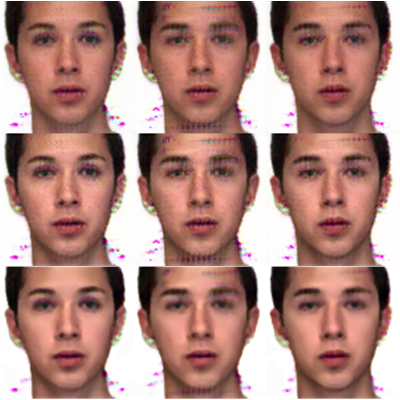}\label{fig:out_syn}}\hfill
	\subfloat[original outputs]
	{\includegraphics[width=3cm]{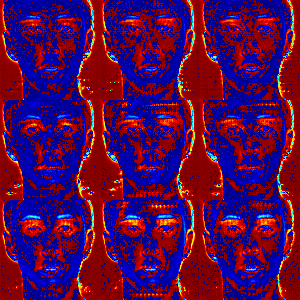}\label{fig:hm1}}\hfill
	\subfloat[after CLAHE]
	{\includegraphics[width=3cm]{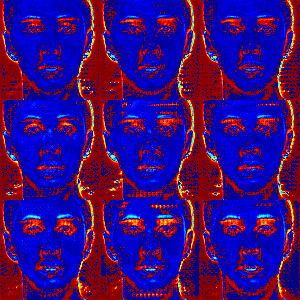}\label{fig:hm2}}\hfill\null
	\caption{(a) Source and target images of the same tester.
		(b) From left to right: syntheses created by two baselines and GATH; 
		top to bottom: raw outputs, histogram-equalized (CLAHE) outputs and
		sharpened outputs, respectively.
		(c,d) The pixel-wise error heat maps of one sample in two cases: the raw output
		and CLAHE-applied output. In each figure, from left to right: output error of GATH-DC,
		GATH-C and GATH, respectively;
		from top to bottom: error heat maps on three channels B, G and R, respectively.}
	\label{fig:er_heat}
\end{figure}

Table~\ref{tab:pixel_er} shows pixel-wise MAE and RMSE when comparing the synthesized output 
with the ground truth image of the same subject (i.e. $x_{tgt}$ and $y_{tgt}$ are of the same person, 
manifesting the same expression and they should look the very similar). 
The errors are organized by dataset, and gathered from four different settings:
whether taking the background error into account or not (by using a mask to localize the face region),
and whether using CLAHE to increase the contrast of the outputs. It is observed that training
only the generator, without feedback from the discriminator and classifier, actually makes the model
produce better pixel-wise color reconstruction without using histogram equalization, 
although the difference in error is rather small. After apply CLAHE, the generated outputs of GATH
have smallest errors.
It is also observed from Table~\ref{tab:pixel_er} that the reconstructed background pixels actually
incur higher error than the face region. In addition, the error heat map in Figure~\ref{fig:er_heat}
indicates that errors on the face region are almost uniform, indicating a constant shift
in the color space.

\begin{table}[!ht]
	\centering
	\caption{Pixel-wise MAE and RMSE of intra-class synthesis.}
	\begin{tabular}{lccccccc}
		\cmidrule{1-4}\cmidrule{6-8}    \multirow{2}[4]{*}{} & \multicolumn{3}{c}{MAE} &       & \multicolumn{3}{c}{RMSE} \\
		\cmidrule{2-4}\cmidrule{6-8}          & \multicolumn{1}{c}{RAVDESS} & \multicolumn{1}{c}{VIDTIMIT} & \multicolumn{1}{c}{All} &       & \multicolumn{1}{c}{RAVDESS} & \multicolumn{1}{c}{VIDTIMIT} & \multicolumn{1}{c}{All} \\
		\midrule
		\multicolumn{8}{c}{full image, without CLAHE} \\
		\midrule
		GATH-DC & \textbf{141.53} & \textbf{112.78} & \textbf{127.16} &       & \textbf{182.31} & \textbf{162.47} & \textbf{172.68} \\
		GATH-C & 144.8 & 119.13 & 131.97 &       & 184.66 & 167.34 & 176.21 \\
		GATH  & 146.46 & 118.53 & 132.5 &       & 185.8 & 167.02 & 176.66 \\
		\midrule
		\multicolumn{8}{c}{full image, with CLAHE} \\
		\midrule
		GATH-DC & 94.49 & \textbf{58.66} & 76.58 &       & 138.62 & \textbf{94.26} & 118.53 \\
		GATH-C & 92.31 & 59.35 & 75.83  &       & 136.86 & 95.23 & 117.89 \\
		GATH  & \textbf{91.65} & 59.66 & \textbf{75.66} &       & \textbf{136.21} & 95.89 & \textbf{117.79} \\
		\midrule
		\multicolumn{8}{c}{mask, without CLAHE} \\
		\midrule
		GATH-DC & \textbf{128.48} & \textbf{122.36} & \textbf{125.42} &       & \textbf{174} & \textbf{170.94} & \textbf{172.47} \\
		GATH-C & 134.78 & 128.53 & 131.66 &       & 178.44 & 175.48 & 176.97 \\
		GATH  & 136.14 & 130.16 & 133.15 &       & 179.48 & 176.78 & 178.13 \\
		\midrule
		\multicolumn{8}{c}{mask, with CLAHE} \\
		\midrule
		GATH-DC & 56.47 & 55.42 & 55.94 &       & 93.42 & 85.49 & 89.54 \\
		GATH-C & 57.37 & 55.28 & 56.32  &       & 94.84 & 85.26 & 90.18 \\
		GATH  & \textbf{55.88} & \textbf{54.97} & \textbf{55.43} &       & \textbf{92.63} & \textbf{84.47} & \textbf{88.64} \\
		\bottomrule
	\end{tabular}%
	\label{tab:pixel_er}%
\end{table}%

\begin{table}[!ht]
	\centering
	\caption{RMSE of Action Unit Itensity in intra-class synthesis.}
	\begin{tabular}{lccc}
		\toprule
		& RAVDESS & VIDTIMIT & All \\
		\midrule
		GATH-DC & 0.592 & 0.35  & 0.486 \\
		GATH-C & 0.591 & 0.336 & 0.481 \\
		GATH  & \textbf{0.585} & \textbf{0.334} & \textbf{0.477} \\
		\bottomrule
	\end{tabular}%
	\label{tab:au_er1}%
\end{table}%

However, color reproduction is only one criterion to measure the quality of expression synthesis.
Our main objective in this paper is to synthesize animation driven by AU coefficients. The AU estimation
errors with respect to the ground truth frame are shown in Table~\ref{tab:au_er1}. It proves
that the output of GATH has higher fidelity than the two baselines, as better images will help 
OpenFace estimate AU intensities more accurately. It is also unsurprising
that GATH-C, trained jointly with a discriminator, performs better than GATH-DC. These results
prove the benefit of our proposed GATH model in synthesizing facial expressions.

\begin{figure}[!ht]
	\centering
	\includegraphics[width=\linewidth]{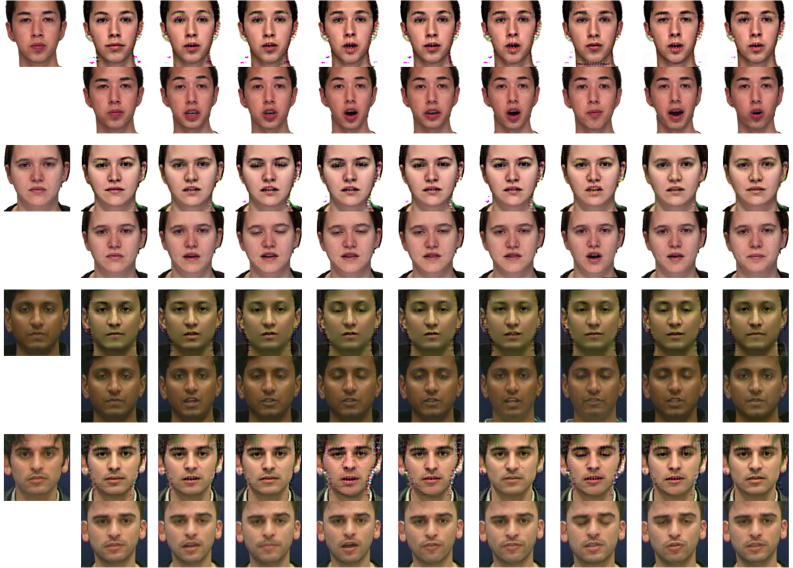}
	\caption{Samples from four sequences in paired evaluation.
	For each sequence, the top left is the still source image $x_{src}$, and it is
	post-processed. $x_{src}$ is at neutral expression.
	In each vertical pair, the top image is the hallucinated frame $x_{tgt}$, while
	the corresponding target frame $y_{tgt}$ is at the bottom. In the $4^{th}$ sequence,
	the source image and the target video were captured at two different occasions.}
	\label{fig:eval_pair}
\end{figure}

\begin{figure}[!ht]
	\centering
	\includegraphics[width=\linewidth]{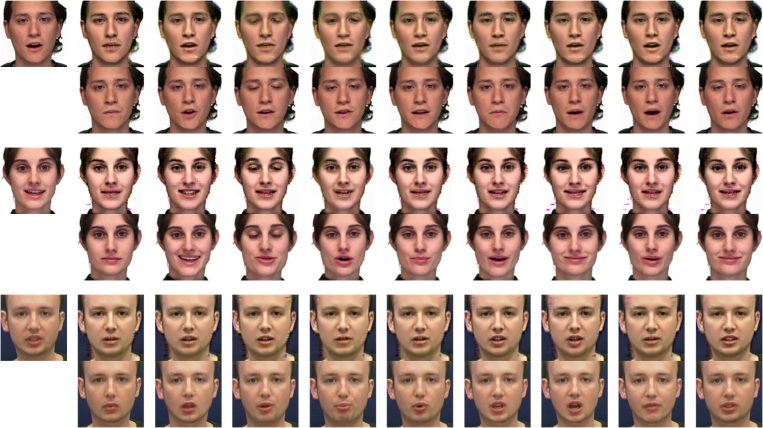}
	\caption{Samples from three sequences, starting with non-neutral expression source images. 
		The model learned to synthesize the "closed lip" expression when target is neutral.}
	\label{fig:eval_pair_exp}
\end{figure}

Figures~\ref{fig:eval_pair},~\ref{fig:eval_pair_exp} demonstrate the synthetic results of GATH. 
The generated texture has been enhanced visually with the aforementioned post-processing procedure.
Notice that in Figure~\ref{fig:eval_pair}b,c,d, it is shown that our model is able to
hallucinate eye-blinking motions.

\subsection{Inter-class Synthesis Evaluation}

\begin{table}[!ht]
	\centering
	\caption{RMSE of Action Unit intensity in inter-class synthesis.}
	\begin{tabular}{ccc}
		\toprule
		GATH-DC & GATH-C & GATH \\
		\midrule
		0.587 & 0.583 & \textbf{0.579} \\
		\bottomrule
	\end{tabular}%
	\label{tab:au_er2}%
\end{table}%

In this evaluation, we compare the AU estimation scores returned by OpenFace on
the ground truth of a subject, and scores on the corresponding syntheses. 
The results are shown in Table~\ref{tab:au_er2}. 
Once again, the full GATH model outperforms the two baselines.

\subsection{Qualitative Assessments}

\begin{figure}[!ht]
	\centering
	\includegraphics[width=\linewidth]{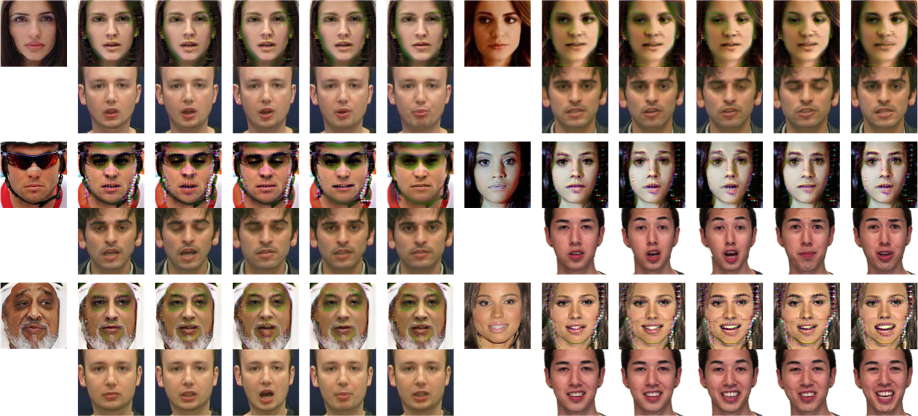}
	\caption{Samples from six animated sequences, with the source images taken randomly
	from CelebA dataset, with different genders, skin colors, hair styles, etc.}
	\label{fig:celeba}
\end{figure}

\begin{figure}[!ht]
	\centering
	\includegraphics[width=\linewidth]{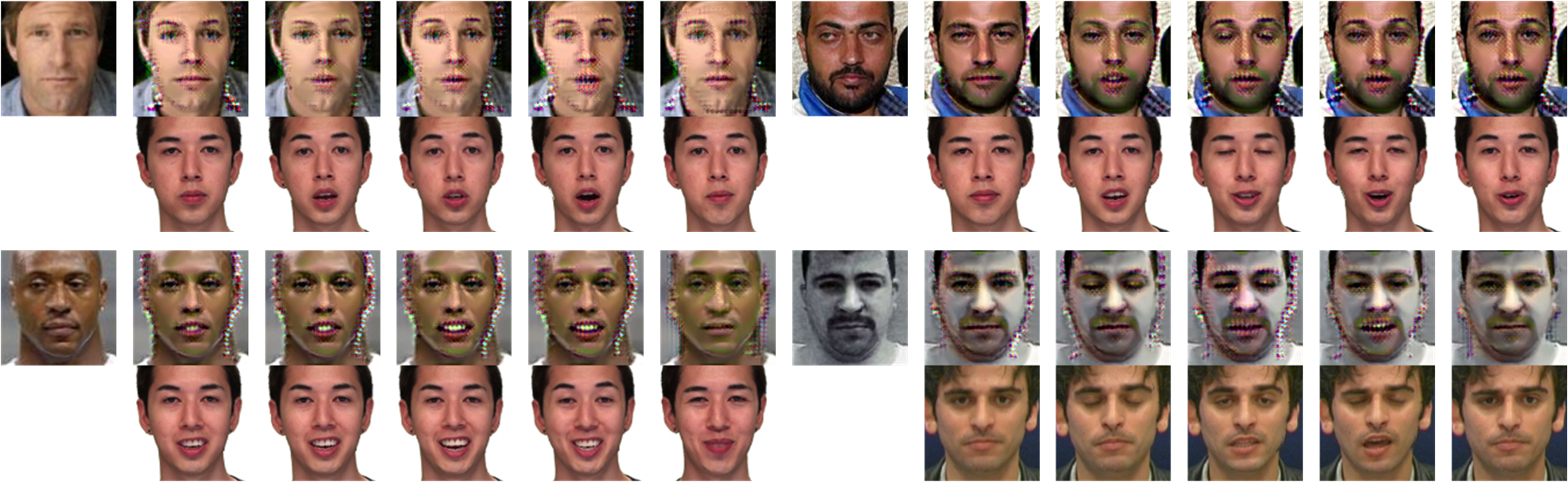}
	\caption{Samples from four animated sequences, in which the source images were taken randomly
		from LFW dataset.}
	\label{fig:lfw}
\end{figure}

Figure~\ref{fig:celeba} and~\ref{fig:lfw} show animated sequences by GATH, in which the source
images are sampled from the CelebA and LFW datasets, respectively, with diversity across
genders, skin colors, styles etc. Interestingly in Figure~\ref{fig:celeba}c, the model even 
synthesizes eyes beyond the shades. More samples are shown in the supplementary
video\footnote{\url{https://www.youtube.com/watch?v=Zr9MlAazPpo}}.

\subsection{Template-and-target-free Expression Editing}

\begin{figure}[!ht]
	\centering
	\includegraphics[width=\linewidth]{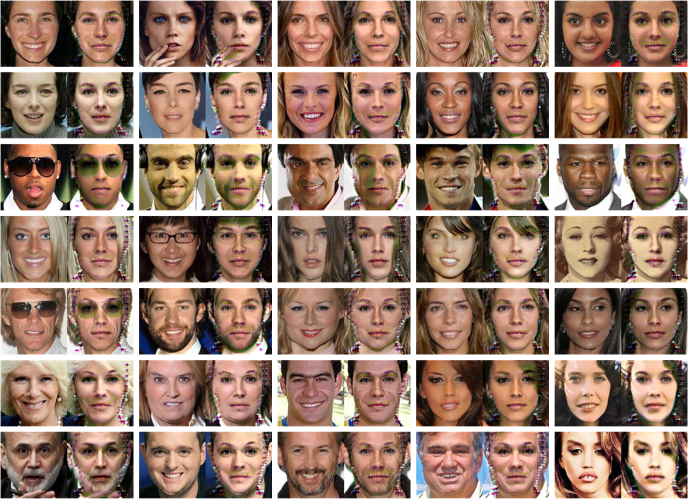}
	\caption{Samples from expression suppression editing with our model on the CelebA dataset.
	In each pair of images, the source is on the left, the suppressed synthesis is shown on the right.}
	\label{fig:editing}
\end{figure}

\begin{figure}[!ht]
	\centering
	\includegraphics[width=\linewidth]{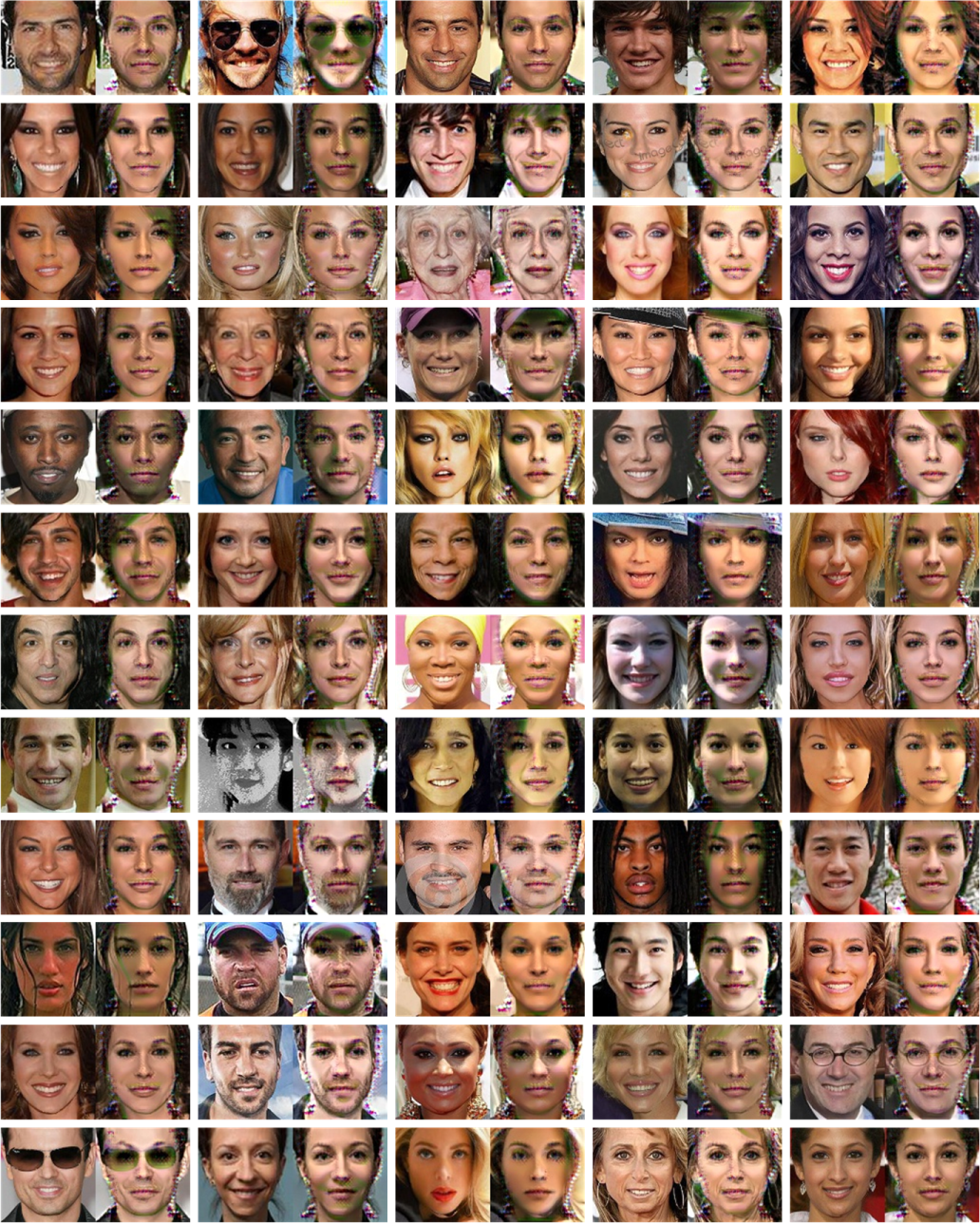}
	\caption{More samples from expression suppression editing with our model on the CelebA dataset.}
	\label{fig:editing2}
\end{figure}

In this experiment, we perform expression suppression, transforming a face with arbitrary
expression back to the neutral pose. Source images are sampled from the CelebA dataset.
The qualitative results are illustrated in Figure~\ref{fig:editing},\ref{fig:editing2}.
We transform the source to neutral expression simply by providing GATH with a zero AU vector.
It proves that via our learning framework, the generator has learned to disentangle the
identity code from expression. Thus, giving GATH zero AU coefficients equals generating
a neutral face of the source actor.

\subsection{Limitations}
Our GATH model has proved to be able to synthesize novel face from arbitrary source. 
However, there still exists some issues remaining:
\begin{itemize}
	\item The synthesized image loses its texture dynamic range.
	\item There is still color noise and distortions in the reconstructed face, especially around
	the face contour and strong edges.
\end{itemize}

We will investigate these issues thoroughly to make GATH more robust and generate higher
quality face synthesis in future work.


\section{Conclusions}

In this paper, we introduce Generative Adversarial Talking Head, a generative neural net
that is capable of synthesizing novel faces from any source portrait given a vector of action unit
coefficients. In our GATH framework, we jointly train a generator with a adversarial discriminator
and a classifier, while being supervised by an AU estimator to make the generator learn
correct expression deformations, as well as disentangle the identity features from expression.
Our model directly manipulates image pixels to hallucinate a novel facial
expression, while preserving the individual characteristics of the source face, without
using a statistical face template or any texture rendering.

\linespread{1.0}

\bibliographystyle{splncs}
\bibliography{mybib}

\end{document}